\documentclass[runningheads]{llncs}
\usepackage[T1]{fontenc}
\usepackage{pifont}
\usepackage{graphicx,verbatim}
\usepackage{medmath}  
\usepackage{multirow}
\usepackage{xcolor} 
\definecolor{darkgreen}{RGB}{0,100,0}
\definecolor{darkred}{RGB}{139,0,0}
\usepackage{hyperref}
\usepackage{color}

\begin{document}
\title{Better Eyes, Better Thoughts: Why Vision Chain-of-Thought Fails in Medicine}
\titlerunning{Better Eyes, Better Thoughts}
%
\author{Yuan Wu\inst{1}\thanks{Equal contribution.} \and
Zongxian Yang\inst{1}\protect\footnotemark[1] \and
Jiayu Qian\inst{1}\protect\footnotemark[1] \and 
Songpan Gao\inst{1} \and 
Guanxing Chen\inst{1} \and 
Qiankun Li\inst{2} \and 
Yu-An Huang\inst{3} \and 
Zhi-An Huang\inst{1}$^\dagger$}
\authorrunning{Y. Wu et al.}
%
\institute{Department of Computer Science, City University of Hong Kong (Dongguan) \and
College of Computing and Data Science (CCDS), Nanyang Technological University \and
School of Computer Science, Northwestern Polytechnical University}


  
\maketitle 
\begingroup
\renewcommand{\thefootnote}{\dag}
\footnotetext{Corresponding author: huang.za@cityu-dg.edu.cn}
\endgroup

%
\begin{abstract}
Large vision-language models (VLMs) often benefit from chain-of-thought (CoT) prompting in general domains, yet its efficacy in medical vision-language tasks remains underexplored. We report a counter-intuitive trend: on medical visual question answering, CoT frequently underperforms direct answering (DirA) across general-purpose and medical-specific models. We attribute this to a \emph{medical perception bottleneck}: subtle, domain-specific cues can weaken visual grounding, and CoT may compound early perceptual uncertainty rather than correct it.
To probe this hypothesis, we introduce two training-free, inference-time grounding interventions: (i) \emph{perception anchoring} via region-of-interest cues and (ii) \emph{description grounding} via high-quality textual guidance. Across multiple benchmarks and model families, these interventions improve accuracy, mitigate CoT degradation, and in several settings reverse the CoT--DirA inversion. Our findings suggest that reliable clinical VLMs require robust visual grounding and cross-modal alignment, beyond extending text-driven reasoning chains. Code is available \href{https://github.com/TianYin123/Better_Eyes_Better_Thoughts}{here}.



\end{abstract}

\section{Introduction}
\label{sec:Int}
Large vision-language models (VLMs) have demonstrated strong capabilities across a wide range of multimodal tasks~\cite{qwen3vltechnicalreport,kimiteam2026kimik25visualagentic}. As these models are increasingly explored for healthcare applications, an important requirement emerges: beyond accurate predictions, medical applications necessitate transparent and clinically meaningful rationales~\cite{liu2024medcot}. Chain-of-thought (CoT) prompting is a natural candidate, as it produces explicit step-by-step explanations and has often improved both interpretability and task performance in general domains such as mathematics and science reasoning~\cite{yang2025med,wei2022chain,guo2025deepseek}. It is therefore tempting to expect similar benefits in medical vision-language tasks~\cite{WanYua_V2TCoT_MICCAI2025}.

However, whether these presumed benefits truly transfer to medical visual domains remains insufficiently validated. In this work, we present a systematic evaluation showing a counter-intuitive trend: instead of improving performance, CoT often degrades accuracy compared with direct answer (DirA) in medical vision question answering (VQA). This pattern consistently appears across multiple clinical benchmarks and is observed in both general-purpose VLMs and medical-domain models even closed source models. These findings suggest that the empirical success of CoT in general domains does not directly transfer to medical visual reasoning.

To understand this failure, we organize our study around three research questions (RQs): \textit{\ding{172}~whether CoT reliably transfers to medical VQA, \ding{173}~what primarily causes its degradation, and \ding{174}~whether the failure can be mitigated without retraining the model}. Our analysis points to a perception-side limitation rather than a reasoning-only deficiency. We refer to this as the \emph{medical perception bottleneck}: medical images often contain subtle lesions and highly specialized visual cues, making accurate visual grounding and cross-modal alignment substantially more difficult than in general-domain settings. When the initial visual grounding is weak or ambiguous, longer CoT generations may propagate and amplify early perceptual errors rather than correct them~\cite{amjith2025can,ye2026unveiling,kang2025hssbench}. Our experiments provide consistent evidence that CoT is more sensitive to visual perception than DirA.

To verify our proposed hypothesis, we introduce two targeted, training-free interventions that operate purely at inference. The first, \emph{perception anchoring}, provides
region of interest (RoI) bounding box to guide visual attention toward clinically relevant regions. The second, \emph{description grounding}, supplements model input with expert-level textual descriptions to better align visual evidence with medical semantics. Experiment results validate the rationality of our hypothesis. These interventions substantially improve performance in our evaluated settings, and the gains are generally more pronounced for CoT than for DirA. Notably, they are sufficient to mitigate the observed performance inversion and allow CoT to recover its advantage.  Because both strategies require no parameter updates, they are particularly compatible with real-world clinical deployment, where access to model weights, large-scale training data, and retraining resources is often limited. The main contributions of this work are summarized as follows:
\begin{itemize}
    \item We provide a systematic empirical study showing that CoT prompting can substantially degrade performance in medical visual question answering, despite its strong performance in general-domain reasoning tasks.
    
    \item We propose the \emph{medical perception bottleneck} hypothesis and present supporting evidence that CoT is especially sensitive to errors in visual grounding and cross-modal alignment.

    \item We introduce two effective, training-free interventions in inference-time, namely perception anchoring and description grounding, which help mitigate CoT degradation without additional model retraining and improve the practical deployability of medical VLMs.
\end{itemize}

\section{Methodology}
\label{methods}

Motivated by the \emph{medical perception bottleneck} discussed in Section~\ref{sec:Int}, this section links our empirical observations with a structural perspective on medical VLM inference. We present a three-stage decomposition of medical VQA reasoning to interpret how imperfect visual grounding may influence CoT generation, and introduce two targeted interventions as controlled probes to examine this effect.
\begin{figure}[t]
    \centering
    \includegraphics[width=1\linewidth]{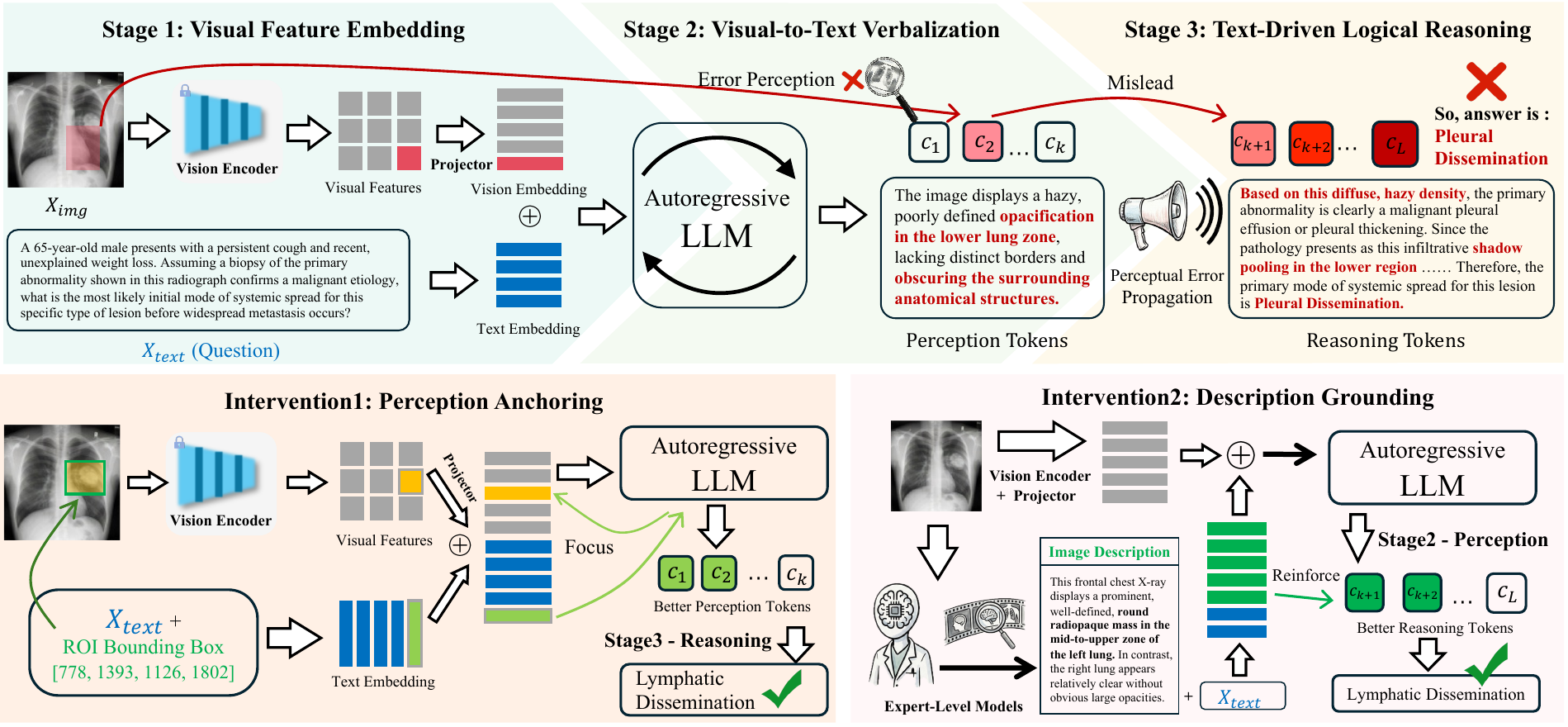}
    \caption{The three-stage Medical VLM CoT framework and targeted interventions. }
    \label{fig:method_main}
\end{figure}

\subsection{A Three-Stage View of Perception Bottlenecks in CoT}
\label{2_1}
To contextualize the medical perception bottleneck, we describe medical VQA inference  in Figure~\ref{fig:method_main} through a three-stage decomposition of VLM-based CoT generation.
Given a medical image $X_{img}$ and a textual query $X_{text}$, let $Z_{vis}$ denote a sequence of visual tokens projected into the LLM embedding space via the vision encoder and projection layer.

\noindent \textbf{Direct Answer.} In DirA mode, the model predicts the final answer $A$ through single-step cross-modal inference:
\begin{equation}
P_{DA}(A) = P(A \mid X_{text}, Z_{vis}).
\end{equation}
When visual evidence is weak or ambiguous, DirA may rely more heavily on textual priors and coarse visual cues, without explicitly verbalizing intermediate perceptual interpretations.

Inspired by previous research~\cite{wang2025unified} on general multi-modal CoT, we decompose the VLM CoT process into three sequential stages to interpret the failure modes observed in medical settings:

\noindent \textbf{Stage 1: Visual Feature Embedding.}
The image and query are represented as a concatenated token sequence in the LLM input space:
\begin{equation}
H_{\text{embed}} = [Z_{vis}; \mathcal{E}(X_{text})],
\end{equation}
where $\mathcal{E}(\cdot)$ denotes the text embedding layer.

\noindent \textbf{Stage 2: Visual-to-Text Verbalization.}
The model first produces perceptual description tokens $C_{\text{perc}} = [c_1, c_2, \dots, c_k]$ that verbalize the visual evidence:
\begin{equation}
P(C_{\text{perc}} \mid H_{\text{embed}}) = \prod_{i=1}^{k} P(c_i \mid H_{\text{embed}}, c_{1:i-1}).
\end{equation}
Due to domain-specific cues and subtle findings in medical images, this verbalization step may compress visual features into coarse or ambiguous textual descriptions, introducing perceptual uncertainty into the generated context.

\noindent\textbf{Stage 3: Text-Driven Reasoning.}
As the reasoning chain extends, prior works suggest that model attention can gradually shift from the original visual tokens to the previously generated text tokens~\cite{amjith2025can,ye2026unveiling}.
Conditioned on $C_{\text{perc}}$, the model then generates reasoning tokens $C_{\text{reason}} = [c_{k+1}, \dots, c_L]$ to derive the answer:
\begin{equation}
P(C_{\text{reason}} \mid H_{\text{embed}}, C_{\text{perc}}) = \prod_{j=k+1}^{L} P(c_j \mid H_{\text{embed}}, C_{\text{perc}}, c_{1:j-1}).
\end{equation}

\noindent \textbf{Perceptual Error Propagation.} This decomposition highlights a key failure pattern in medical CoT: when perceptual uncertainty or inaccuracies are introduced during Stage 2, subsequent Stage 3 reasoning may increasingly condition on the generated textual context rather than the original visual evidence, causing the reasoning trajectory to drift and accumulate deviations from the image-grounded facts~\cite{kang2025hssbench}.

\subsection{Bridging the Modality Gap: Spatial and Semantic Interventions}
\label{Interventions}
To verify the existence of the perception bottleneck, we propose two approaches to restore visual perception before Stage 3. If our hypothesis holds, bridging this gap should recover the medical CoT performance of VLMs.

\noindent\textbf{Intervention 1: Perception Anchoring via RoI.}
Previous studies have demonstrated that introducing RoI bounding boxes can help the model achieve better perception~\cite{chen2025thinktwicemoreiterative,Du_2022_CVPR}.
Instead of modifying model parameters or visual features, we implement perception anchoring at the prompt level.
RoI information is tokenized and concatenated along the sequence dimension to form an anchored input context:
\begin{equation}
H_{\text{input}}^{\text{anchored}} = \left[ Z_{vis}; \mathcal{E}([X_{text}; X_{RoI}]) \right].
\end{equation}
This provides explicit spatial priors that help focus attention on clinically relevant regions during Stage~2 verbalization, reducing perceptual ambiguity and encouraging more visually grounded CoT reasoning.

\noindent\textbf{Intervention 2: Expert-level Description Grounding.}
Following previous studies~\cite{zheng2023ddcot,wang2025unified}, we introduce explicit medical descriptions $T_{expert}$ generated by an expert-level model to examine whether improved visual textualization can stabilize reasoning.
The generation process is conditioned on both visual tokens and grounded textual cues:
\begin{equation}
P(c_t \mid X_{text}, Z_{vis}, T_{expert}, c_{1:t-1}).
\end{equation}
By supplying high-quality semantic guidance, this intervention acts as an oracle-style probe that reduces perceptual uncertainty during visual-to-text verbalization, allowing the model’s latent reasoning capabilities to be effectively utilized.

\section{Experiments}
To systematically evaluate our hypothesis, this section is organized around three RQs. 
First, we examine whether CoT's general-domain gains transfer to medical VQA. 
Next, we probe whether CoT is disproportionately sensitive to degraded or counterfactual visual inputs, which would be consistent with a perception-side bottleneck. 
Finally, we evaluate whether inference-time grounding interventions can restore CoT performance by bridging the perception gap.

\subsection{Experimental Setup}
\subsubsection{Benchmarks and Models.}
We conducted experiments on five medical benchmarks: VQA-RAD~\cite{lau2018dataset}, SLAKE~\cite{liu2021slake}, PMC-VQA~\cite{zhang2023pmc}, Path-VQA~\cite{he2020pathvqa}, OmniMedVQA~\cite{hu2024omnimedvqa}; 
and five general benchmarks: CVBench~\cite{tong2024cambrian}, BLINK~\cite{fu2024blink}, ScienceQA~\cite{saikh2022scienceqa}, AI2D~\cite{kembhavi2016diagram}, MathVista~\cite{lu2023mathvista}. 
These benchmarks include multiple-choice, closed-ended, and open-ended question formats. 
To test generality, we evaluated closed-source models (Gemini-3-Flash, GPT-4o-mini, Grok-4-Fast), open-source general-purpose models (Qwen3-VL-Instruct~\cite{qwen3vltechnicalreport}, Intern-VL3~\cite{zhu2025internvl3}), and medical domain-specific models (Lingshu~\cite{xu2025Lingshu}, Hulu-Med~\cite{jiang2025hulu}).

\begin{figure}[t]
    \centering
    \includegraphics[width=1\linewidth]{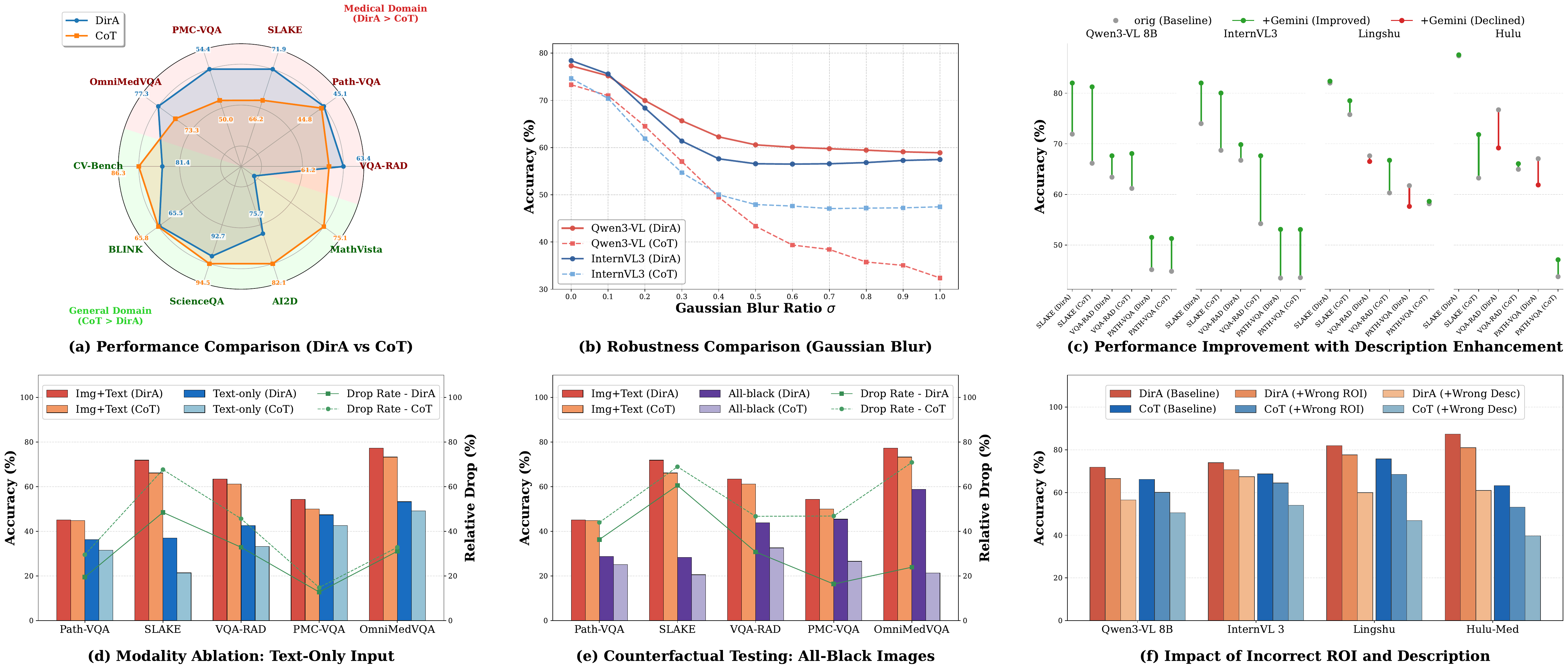}
    \caption{Main results across RQ1--RQ3. (a) CoT improves general benchmarks but degrades medical benchmarks. 
    (b) CoT is more sensitive to progressive visual degradation than DirA. 
    (c) Supplementing models with expert-level image descriptions alone effectively mitigates CoT degradation.
    (d,e) Counterfactual inputs reveal pseudo-robustness in DirA and stronger visual dependence in CoT.
    (f) Incorrect RoI and descriptions degrade CoT performance, confirming its reliance on accurate visual grounding.}
    \label{fig:mainResult}
\end{figure}

\subsubsection{Implementation Details.}
Following the evaluation setting in Hulu-Med~\cite{jiang2025hulu}, we utilize the vLLM framework~\cite{vllm} for model inference, with the maximum generation length capped at 8,192 tokens.All results are averaged over three inference runs. 
To distinguish DirA from CoT, we use the prompts ``Answer the question using a single word or phrase.'' and 
``Please reason step by step, and put your final answer within \textbackslash boxed\{\}.'' respectively. 

\subsection{\small {Does the success of CoT in general tasks transfer to medical VQA?}}
Figure~\ref{fig:mainResult}a compares Qwen3-VL-Instruct under DirA and CoT. 
CoT consistently improves accuracy on non-medical benchmarks. 
However, across all evaluated medical datasets, introducing intermediate reasoning steps via CoT consistently reduces accuracy compared to DirA.

We further evaluate multiple models on medical benchmarks under DirA and CoT. 
As shown in Table~\ref{tab1}, CoT underperforms DirA across general, medical-specific, and closed-source models. 
This pervasive degradation is consistent with a perception-side bottleneck rather than missing domain knowledge.

\begin{table}[t]
\centering
\caption{Quantitative comparison of DirA and CoT on medical benchmarks. All values report accuracy (\%). Superscripts $^\ast$, $^\dagger$, and $^\ddagger$ denote general, medical-specific, and closed-source models, respectively. Bold indicates the better score between DirA and CoT for each model-dataset pair.}
\label{tab1}
\renewcommand{\arraystretch}{1.3}
\resizebox{\textwidth}{!}{
\begin{tabular}{ll ccccc c}
\hline
Model &  & Path-VQA & SLAKE & VQA-RAD & PMC-VQA & OM.VQA & Avg \\
\hline
\multirow{2}{*}{\shortstack{Qwen3VL\\-8B$^\ast$}} 
    & DirA & \bfseries 45.15 & \bfseries 71.91 & \bfseries 63.41 & \bfseries 54.38 & \bfseries 77.31 & \bfseries 62.43 \\
    & CoT  & 44.82 & 66.16 & 61.19 & 50.01 & 73.31 & 59.10 \\
\hline
\multirow{2}{*}{\shortstack{InternVL3\\-8B$^\ast$}} 
    & DirA & 43.49 & \bfseries 73.99 & \bfseries 66.74 & \bfseries 53.21 & \bfseries 78.41 & \bfseries 63.17 \\
    & CoT  & \bfseries 43.57 & 68.71 & 54.23 & 49.84 & 74.62 & 58.19 \\
\hline
\multirow{2}{*}{\shortstack{Lingshu\\-7B$^\dagger$}} 
    & DirA & \bfseries 61.73 & \bfseries 82.00 & \bfseries 67.62 & \bfseries 54.31 & \bfseries 86.28 & \bfseries 70.41 \\
    & CoT  & 58.16 & 75.77 & 60.31 & 47.58 & 67.54 & 61.87 \\
\hline
\multirow{2}{*}{\shortstack{Hulu-Med\\-7B$^\dagger$}} 
    & DirA & \bfseries 67.06 & \bfseries 87.37 & \bfseries 76.71 & \bfseries 66.79 & \bfseries 84.16 & \bfseries 76.42 \\
    & CoT  & 43.76 & 63.24 & 64.96 & 59.24 & 79.23 & 62.09 \\
\hline
\multirow{2}{*}{\shortstack{Gemini-3\\-Flash$^\ddagger$}} 
    & DirA & \bfseries 69.94 & \bfseries 84.17 & \bfseries 76.50 & \bfseries 55.48 & \bfseries 74.57 & \bfseries 72.13 \\
    & CoT  & 66.51 & 81.55 & 72.28 & 41.25 & 73.62 & 67.04 \\
\hline
\multirow{2}{*}{\shortstack{GPT-4o\\-mini$^\ddagger$}} 
    & DirA & \bfseries 43.63 & \bfseries 64.25 & \bfseries 50.12 & \bfseries 36.75 & \bfseries 57.61 & \bfseries 50.47 \\
    & CoT  & 42.74 & 61.42 & 45.43 & 31.17 & 50.34 & 46.22 \\
\hline
\multirow{2}{*}{\shortstack{Grok-4\\-fast$^\ddagger$}} 
    & DirA & \bfseries 55.51 & \bfseries 71.80 & \bfseries 64.93 & \bfseries 42.09 & \bfseries 57.72 & \bfseries 58.41 \\
    & CoT  & 53.61 & 70.75 & 62.49 & 41.24 & 57.03 & 57.02 \\
\hline
\end{tabular}}
\end{table}

\subsection{Is CoT reasoning critically bounded by visual perception?}
To examine the perception bottleneck described in Section~\ref{2_1}, we conduct a sensitivity analysis by corrupting visual inputs.
We apply Gaussian blur $\mathcal{G}$ with severity ratios $\sigma \in [0, 1.0]$ to images in OmniMedVQA, evaluated on Qwen3-VL-Instruct 8B and InternVL3 8B.

As shown in Figure~\ref{fig:mainResult}b, CoT performance is substantially more sensitive to progressive visual degradation than DirA.
While DirA degrades more gradually and maintains a non-trivial baseline under severe blur, CoT accuracy drops sharply as visual evidence becomes unreliable.
This asymmetric decline is consistent with our three-stage view: ambiguous visual grounding in Stage~2 can lead to less stable downstream reasoning in Stage~3.

We observe similar patterns under counterfactual inputs (Figure~\ref{fig:mainResult}d,e). 
When removing the image or providing an all-black image, CoT exhibits a larger performance drop than DirA, which is consistent with CoT’s stronger dependence on explicit visual verbalization.
In contrast, DirA maintaining a relatively high baseline under counterfactual inputs suggests a form of pseudo-robustness: higher scores do not necessarily indicate reliable visual grounding, but may reflect heavier reliance on textual priors when visual evidence is unavailable.

\subsection{Can bridging the perception gap reactivate the reasoning potential of medical VLMs?}

\begin{table}[t]
\centering
\caption{Targeted interventions on the SLAKE dataset. Bold indicates the best score within each row (DirA or CoT) across intervention settings.}
\label{tab2}
\resizebox{\textwidth}{!}{
\begin{tabular}{lccccc}
\hline
Model &  & Baseline & + BBox RoI & + Expert Desc. & + Combined Prior\\
\hline
\multirow{2}{*}{\shortstack{Qwen3VL\\-8B}}
    & DirA & 71.91 & 77.95 \fontsize{8pt}{9.6pt}(\textcolor{darkgreen}{+6.04}) & 82.00 \fontsize{8pt}{9.6pt}(\textcolor{darkgreen}{+10.09}) & 84.35 \fontsize{8pt}{9.6pt}(\textcolor{darkgreen}{+12.44}) \\
    & CoT  & 66.16 & 76.15 \fontsize{8pt}{9.6pt}(\textcolor{darkgreen}{+9.99}) & 81.24 \fontsize{8pt}{9.6pt}(\textcolor{darkgreen}{+15.08}) & \textbf{84.45} \fontsize{8pt}{9.6pt}(\textcolor{darkgreen}{+18.29}) \\
\hline
\multirow{2}{*}{\shortstack{InternVL3\\-8B}}
    & DirA & 73.99 & 78.29 \fontsize{8pt}{9.6pt}(\textcolor{darkgreen}{+4.30}) & 82.00 \fontsize{8pt}{9.6pt}(\textcolor{darkgreen}{+8.01}) & 85.11 \fontsize{8pt}{9.6pt}(\textcolor{darkgreen}{+11.12}) \\
    & CoT  & 68.71 & 75.68 \fontsize{8pt}{9.6pt}(\textcolor{darkgreen}{+6.97}) & 80.02 \fontsize{8pt}{9.6pt}(\textcolor{darkgreen}{+11.31}) & \textbf{85.29} \fontsize{8pt}{9.6pt}(\textcolor{darkgreen}{+16.58}) \\
\hline
\multirow{2}{*}{\shortstack{Lingshu\\-7B}}
    & DirA & 82.00 & 84.07 \fontsize{8pt}{9.6pt}(\textcolor{darkgreen}{+2.07}) & 82.37 \fontsize{8pt}{9.6pt}(\textcolor{darkgreen}{+0.37}) & 85.57 \fontsize{8pt}{9.6pt}(\textcolor{darkgreen}{+3.57}) \\
    & CoT  & 75.77 & 82.75 \fontsize{8pt}{9.6pt}(\textcolor{darkgreen}{+6.98}) & 78.51 \fontsize{8pt}{9.6pt}(\textcolor{darkgreen}{+2.74}) & \textbf{85.77} \fontsize{8pt}{9.6pt}(\textcolor{darkgreen}{+10.00}) \\
\hline
\multirow{2}{*}{\shortstack{Hulu-Med\\-7B}}
    & DirA & 87.37 & 87.28 \fontsize{8pt}{9.6pt}(\textcolor{darkred}{-0.09}) & 87.56 \fontsize{8pt}{9.6pt}(\textcolor{darkgreen}{+0.19}) & \textbf{88.50} \fontsize{8pt}{9.6pt}(\textcolor{darkgreen}{+1.13}) \\
    & CoT  & 63.24 & 72.29 \fontsize{8pt}{9.6pt}(\textcolor{darkgreen}{+9.05}) & 71.82 \fontsize{8pt}{9.6pt}(\textcolor{darkgreen}{+8.58}) & 77.85 \fontsize{8pt}{9.6pt}(\textcolor{darkgreen}{+14.61}) \\
\hline
\end{tabular}}
\end{table}

Building upon Section~\ref{methods} and the sensitivity analysis above, we hypothesize that if visual perception is the primary bottleneck for CoT failures in medical VLMs, then bridging this gap should disproportionately restore CoT accuracy.
We evaluate two training-free interventions: perception anchoring (+ BBox RoI) and expert-level description grounding (+ Expert Desc.). 
Table~\ref{tab2} reports results on SLAKE (the only benchmark providing RoI bounding boxes).

Overall, grounding interventions systematically restore CoT performance and often benefit CoT   than DirA. Perception anchoring yields consistent gains, with CoT exhibiting higher sensitivity to localization priors than DirA. Description grounding alone further improves CoT (Figure~\ref{fig:mainResult}c), suggesting that improved visual textualization can successfully stabilize downstream reasoning. Conversely, injecting incorrect RoI and descriptions severely degrades performance (Figure~\ref{fig:mainResult}f), confirming from a counterfactual perspective that CoT's reasoning chain is critically dependent on accurate initial visual grounding. Most notably, combining both interventions reverses the performance inversion for Qwen3VL-8B, InternVL3-8B, and Lingshu-7B, enabling CoT accuracy to match or surpass DirA. 
For Hulu-Med-7B, CoT improves substantially under the combined prior (+14.61\%), but remains below its strong DirA baseline, suggesting that DirA in medical-specialized models may be more influenced by domain language priors.
\begin{figure}[t]
    \centering
    \includegraphics[width=1\linewidth]{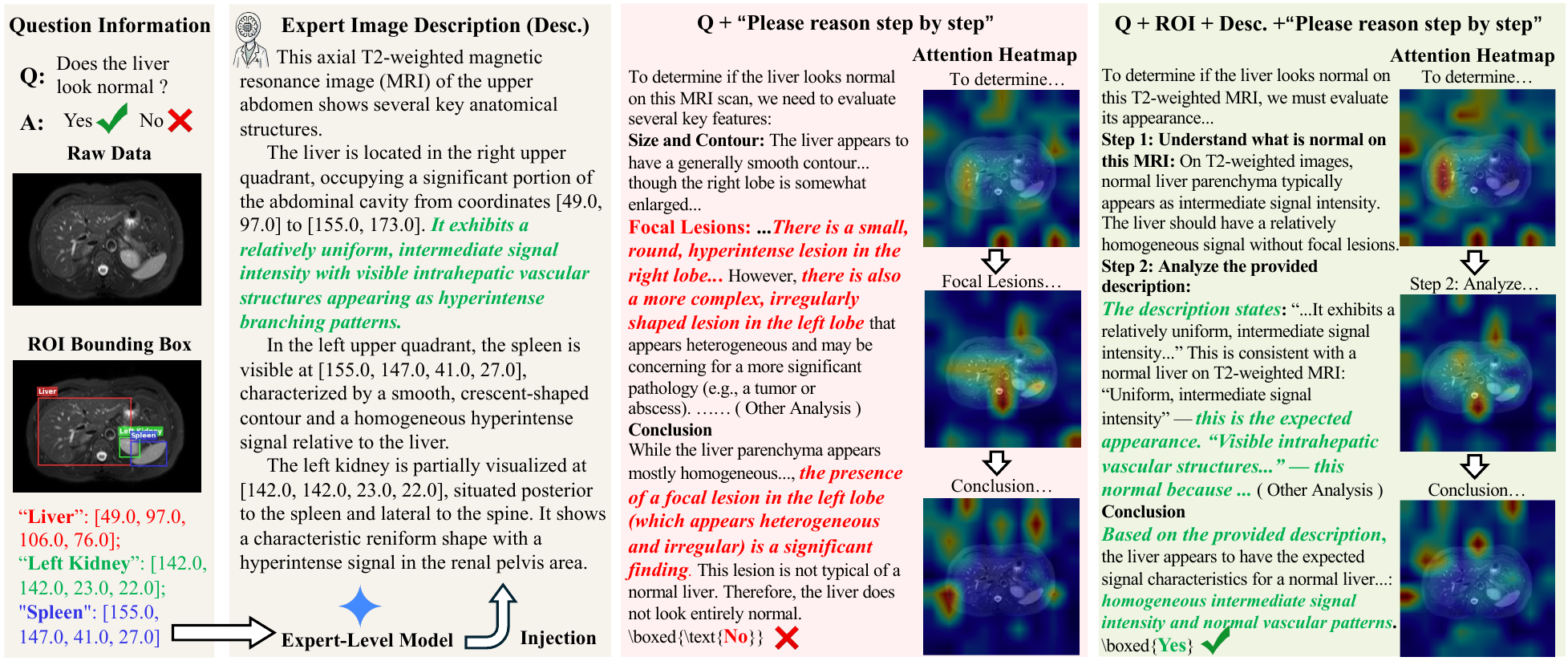}
    \caption{Qualitative case study. Standard CoT exhibits misaligned attention patterns and incorrect conclusions, while grounded interventions provide additional spatial and semantic priors that yield more visually consistent reasoning trajectories.}
    \label{fig:CaseStudy}
\end{figure}

Figure~\ref{fig:CaseStudy} provides a qualitative example. 
Under standard CoT, the model produces an incorrect conclusion accompanied by visually misaligned attention.
After introducing grounding priors (RoI and expert descriptions), the reasoning becomes more consistent with the provided visual/textual evidence, and the attention patterns shift accordingly.

\section{Conclusion and Discussion}
In this work, we report a counter-intuitive trend in medical VLM reasoning: CoT often underperforms direct answering in medical VQA. We attribute this gap to a \emph{medical perception bottleneck}, where weak visual grounding can propagate into downstream CoT generation and destabilize the reasoning trajectory. Through targeted, training-free interventions, we show that these models' reasoning ability is not necessarily absent, but can be substantially improved when perceptual grounding is strengthened.

Beyond improving benchmark performance, our findings carry practical implications for clinical deployment. Medical images rarely appear in isolation; they are typically accompanied by structured reports, radiologist notes, and referral indications. Our results suggest that providing VLMs with spatial and semantic grounding cues, which can be derived from existing clinical documentation or lightweight localization priors, may offer a pragmatic pathway to more reliable on-premise clinical AI assistants without retraining large models. Overall, we argue that future progress should prioritize bridging the vision and language grounding gap, rather than solely extending text-driven reasoning chains.

%
%
%
\bibliographystyle{splncs04_custom}
\bibliography{mybibliography}

\end{document}